# Autoencoder-based prediction of ICU clinical codes

Tsvetan R. Yordanov[1,2][0000-0003-0554-6646], Ameen Abu-Hanna[1,2][0000-0003-4324-7954],
Anita CJ Ravelli[1,2][0000-0002-3447-8286], Iacopo Vagliano[1,2][0000-0002-3066-9464]

[1] Amsterdam UMC, University of Amsterdam, dept Medical informatics,
[2] Amsterdam Public Health research institute, Meibergdreef 9, Amsterdam, Netherlands
```
{t.yordanov,a.abu-hanna,
a.c.ravelli,i.vagliano}@amsterdamumc.nl
```

**Abstract.** Availability of diagnostic codes in Electronic Health Records (EHRs) is crucial for patient care as well as reimbursement purposes. However, entering them in the EHR is tedious, and some clinical codes may be overlooked. Given an incomplete list of clinical codes, we investigate the performance of ML methods on predicting the complete ones, and assess the added predictive value of including other clinical patient data in this task. We used the MIMIC-III dataset and frame the task of completing the clinical codes as a recommendation problem. We consider various autoencoder approaches plus two strong baselines; item co-occurrence and Singular Value Decomposition (SVD). Inputs are 1) a record's known clinical codes, 2) the codes plus variables. The co-occurrence-based approach performed slightly better (F1 score=0.26, Mean Average Precision [MAP]=0.19) than the SVD (F1=0.24, MAP=0.18). However, the adversarial autoencoder achieved the best performance when using the codes plus variables (F1=0.32, MAP=0.25). Adversarial autoencoders performed best in terms of F1 and were equal to vanilla and denoising autoencoders in term of MAP. Using clinical variables in addition to the incomplete codes list, improves the predictive performance of the models.

## 1 Introduction

During a patient's stay at an Intensive Care Unit (ICU) their record will often have multiple diagnostic and procedure codes entered. The clinical codes for a patient may be incomplete or missing for various reasons. To help automate data entry, we investigate the task of predicting clinical diagnostic and procedure codes for ICU patients. We consider missing clinical codes as a recommendation problem, and develop a Recommender System (RS) that suggests clinical codes for a patient stay given what is already present in their EHR record. We focus on AutoEncoders (AEs) as they are known to perform well at recommendation problems [1, 2]. In the current experiments we investigate different AE types and apply them to the problem of clinical code recommendation.

We aimed to evaluate the predictive performance of different types of AEs (vanilla, denoising, variational, and adversarial) using two types of input data: 1. the known clinical codes for a record, and 2. the codes and variables [structured data]). We compare their results against two baseline models.

## 2    Methods

**Problem statement.** We define a set of $m$ ICU patient records $P$ and a set of $n$ clinical codes $C$, where we attempt to model the spanned space of $P \times C$. A sparse matrix $X \in \{0, 1\}^{m \times n}$ is used to represent the codes of each patient ICU stay where $X_{ij}$ indicates the presence of clinical code $j$ in patient record $i$. Additional information from a record is held in the supplementary information matrix $S$ [2]. As supplementary information we considered the measurements of patient vitals during ICU stay (e.g., heart rate) along with demographics (e.g., age).

**Dataset.** The MIMIC-III dataset is a large freely-available dataset of 57,786 ICU hospital admissions of 46,476 patients from one US hospital between 2001-2012 [3].

For structured data, each record holds laboratory measurements and vital signs, patient demographics, and a list of clinical diagnostic and procedure codes. Additionally, there are semi-structured data like the textual descriptions of the clinical codes, and unstructured data like clinical notes.

The International Statistical Classification of Diseases and Related Health Problems version 9 (ICD9) was used for coding patient conditions and the procedures received. ICD9 consists of about 13,000 unique codes in varying levels of detail.

**Inclusion/Exclusion criteria.** We consider codes which appear at least 50 times in the dataset [4] and only include admissions of patients older than 18 years of age.

**Features.** We used the list of known diagnostic and procedure codes per patient ICU stay. For the various AEs we also included the supplementary information of the structured patient data.

We considered all numeric and categorical features available during the whole ICU stay. Time-series were aggregated with the mean and the difference between the last and first measurement. Time-series were imputed by a sliding-window average of length one, where a missing measurement was set to the average of the two non-missing nearest measurements.

**Models.** We compare performances between four types of autoencoders – 1) vanilla AEs, 2) Denoising Autoencoders (DAEs) [5], 3) Variational Autoencoders (VAEs) [6], and 4) Adversarial Autoencoders (AAEs) [7]. We evaluated each AE type using the two different input sets described earlier.

We considered the following two baseline models: 1) a simple co-occurrence based method and 2) the Singular Value Decomposition (SVD) matrix factorization method. Since these baselines do not support the provision of supplementary information, only the lists of ICD codes were used as input.

**Model training and evaluation strategy.** We follow the pre-processing steps described in [4]. We used 5-fold Cross-Validation (CV) with a 80/10/10 split of each fold into training ($P_{train}$), validation ($P_{val}$) and test sets ($P_{test}$). Hyperparameter values obtained from using grid-search in the first fold were then used and not re-computed for the remaining folds.

During training, models were provided with the full list of clinical codes, and where applicable supplementary data, from a patient record, while during evaluation on $P_{val}$ or $P_{test}$ they received a randomly-sampled 50% subset of the codes. The models were evaluated on predicting the missing codes.

For tuning the count-based and SVD models, we used correlation order and number of dimensions, respectively. For AEs, we optimized the learning rate, number of epochs, batch size, number of neurons per hidden layer, and for the latent space.

**Performance Metrics.** We report on the F1 score for measuring the harmonic mean of the precision and recall of the recommendations, as well as the Mean Average Precision (MAP) for measuring the relevance of the recommendations.

## 3 Results

**Table 1.** Averaged performance metric results from 5-fold cross-validation on the test sets.

|  |  | only codes | codes + TD |
|---|---|---|---|
| Item co-occurrence | F1 (SD) | 0.26 (0.0019) | - |
|  | MAP (SD) | 0.19 (0.0018) | - |
| SVD | F1 (SD) | 0.24 (0.0014) | - |
|  | MAP (SD) | 0.18 (0.0012) | - |
| Vanilla AE | F1 (SD) | 0.31 (0.0035) | 0.31 (0.0034) |
|  | MAP (SD) | 0.25 (0.0037) | 0.25 (0.0035) |
| DAE | F1 (SD) | 0.30 (0.0035) | 0.31 (0.0039) |
|  | MAP (SD) | 0.24 (0.0034) | 0.25 (0.0043) |
| VAE | F1 (SD) | 0.20 (0.0029) | 0.23 (0.0033) |
|  | MAP (SD) | 0.13 (0.0025) | 0.16 (0.0026) |
| AAE | F1 (SD) | 0.31 (0.0059) | **0.32 (0.0013)** |
|  | MAP (SD) | 0.25 (0.0055) | 0.25 (0.0019) |

Types of input data considered: only clinical codes ('**only codes**'); clinical codes and Tabular Data ('**codes + TD**'). SVD – Singular Value Decomposition; DAE – Denoising Autoencoder; VAE - Variational Autoencoder; AAE – Adversarial Autoencoder.

**Dataset.** The pre-processed dataset contained 49,002 admissions of 38,402 patients. 822,047 codes were recorded with more than 50 occurrences, of which 1,581 unique codes (1,208 diagnostic and 372 procedural). The median number of codes given per record was 15 with an Inter-Quartile-Range (IQR) of 10-21.

**Performance.** All AE's apart from VAEs achieved a higher F1 score and MAP compared to the baselines (**Table 1**). The count-based predictor achieved a slightly higher F1 (0.26) than the SVD (0.24), as well as a slightly better MAP (0.19 versus 0.18). Overall, the best model was the AAE using supplementary data (F1 0.32; MAP 0.25), followed by the vanilla AE and DAE using the same supplementary information (F1 0.31; MAP 0.25).

## 4      Discussion

In our experiments we demonstrated that for an ICU patient record using supplementary information in addition to the incomplete clinical code list leads to an improvement in the F1 score and MAP metric in 3 out of 4 and 2 out of 4 types of AEs respectively.

**Related work.** The challenge of predicting clinical codes in ICU EHR patients has received attention in recent works [8, 9]. In all comparative literature the authors formulated the problem as a Multilabel Classification (MC) task, and thus their performance metrics reported should not be directly compared to those from our experiments where we phrased the problem as a RS task.

Xu et al. developed an ensemble ML method for predicting ICD codes in the MIMIC-III dataset [9]. They used the 32 most frequently-occurring ICD codes. Their ensemble model achieved an F1-micro of 0.76, but that work addressed a simpler problem than ours by using only the most common ICD codes, and the architecture of the proposed solution involved a rather complex ensemble of ML methods.

Bao et al. [8] developed a hybrid capsule network model om MIMIC-III using only the clinical notes as inputs. Their best performance F1-micro was 0.67. Unlike our approach, Bao et al. used rolled-up representation of the clinical codes up to the first three characters and only a selection of 344 codes of three characters for their experiments.

**Strengths and limitations.** Strengths of the current study include its use of a large well-known freely-available EHR dataset, the adoption of pre-processing steps described from previous studies, the employment of cross-validation for optimism correction in the results, and the battery of different types of autoencoders considered. Compared to the previous approaches for clinical code prediction, we adopt a re-phrasing of the task as a recommender system problem and propose a novel application using autoencoders to solve it. Our code is available at https://github.com/tsryo/aae-recommender.

Our study has some limitations. Due to resource constraints, hyperparameter optimization was only done for the first fold of each cross-fold validation and the selected parameters were used in subsequent folds. We only considered autoencoders in the current experiments as they were promising for recommendation tasks. It is unclear whether other deep learning approaches could have performed even better.

**Future work.** In the current approach we did not use semi-structured or unstructured data as inputs, where we suspect there could be further gains in performance possible. Future work should investigate the possibility of incorporating additional sources of information to the ones presented here and report on their impact on model performance. We reported on overall performance across all clinical diagnostic and procedure codes, whilst the question of how that performance would translate into subgroups of codes and code hierarchies was left open. It would be interesting to see how performance varies between the different categories of clinical codes.

We only considered one value for the proportion of codes to hide from a record for evaluating the models' performance (50%). It would be valuable to investigate its impact on model performance.

We excluded clinical codes that occurred less than 50 times in the dataset. Future research should evaluate the impact on performance when including rare codes.

The small gain in performance from using structured supplementary data could potentially increase with more sophisticated encoding techniques, such as using Long-Short Term Memory network [10] or Time2Vec [11] to encode time series, and merits further investigation.

Our system could be used by EHR administrators when performing clinical code entry- where they obtain recommendations for which codes to enter in a record in order to decrease manual workload for instance in case of an applying for reimbursement. Before that can happen, however the utility of introducing such a tool in practice must be evaluated.

## Appendix

## Autoencoders Background

AutoEncoders (AEs) are a family of neural networks that are tasked to reconstruct their inputs. They achieve this by - first encoding their inputs into a compact representation in some latent space $Z$, and then decoding the representations back to the original space. The architecture of a pure vanilla AE consists of an encoder and a decoder, with an information bottleneck in the (midway) hidden layer between them. Specifically, this bottleneck forces the network to learn a more compact representation of its input, hopefully capturing important features and patterns in the data that could then generalize to new inputs unseen by the model before. Such vanilla AE architecture has received several modifications that tried to improve on one or more aspects of its performance.

One such modification is the Denoising AutoEncoder (DAE) which could be trained to perform the task of removing noise artifacts from its inputs [1]. DAEs introduced the concept of corrupting the input to the network's encoder and then training the system to produce an uncorrupted reconstruction.

Another extension of AEs are the so-called Variational AutoEncoders (VAEs), which perform regularization on the latent space $Z$ [2]. To achieve this regularization, VAEs encode their inputs as normal distributions (as opposed to single point-values) and sample from them, later penalizing the network for distributions that differ from the standard normal distribution (in addition to penalizing only on the reconstruction loss) [2].

Finally, inspired from Generative Adversarial Networks (GANs), Adversarial AutoEncoders (AAEs) [3] extend the capabilities of VAEs by allowing one to specify a non-normal prior distribution. AAEs achieve this by removing the need to penalize the network based on a divergence metric (that is used in VAEs), but instead add one additional component to the architecture, a discriminator.

Although they perform unsupervised learning by default, AEs can also be made to perform supervised or semi-supervised learning tasks by performing conditioning on the latent space [4]. Conditioning means embedding into the encoder's output additional labelled information that was not part of the input, for example adding a numeric indicator for a patient's age.

**Supplementary Table 2.** Summary statistics of 49,002 patient demographics in records from the MIMIC-III dataset used in the current study.

| Demographic | Median/ Count | IQR/ Percentage |
|---|---|---|
| Age (years) | 66 | 53-78 |
| LOS (days) | 6,9 | 4-12 |
| Gender (Female) | 21,459 | 43.8% |
| Ethnicity | | |
| White | 35,080 | 71.6% |
| Black | 4,676 | 9.5% |
| Hispanic | 1,679 | 3.4% |
| Asian | 1,134 | 2.3% |
| Other | 1,302 | 2.6% |
| Unknown | 5,131 | 10.5% |
| Admission type | | |
| Emergency | 40,821 | 83.3% |
| Urgent | 1,221 | 2.5% |
| Elective | 6,960 | 14% |

**Supplementary Table 3.** Summary statistics of patient vital signs Time-series measurements in records from the MIMIC-III dataset used in the current study.

| Vital sign | Median | IQR | Number of measurements | Missing % |
|---|---|---|---|---|
| Heart rate (BPM) | 92 | 78-119 | 7,931,088 | 0.2% |
| Respiratory rate (BPM) | 20 | 16-24 | 6,272,372 | 0.3% |
| SpO2 | 98 | 96-99 | 6,079,970 | 0.3% |
| Systolic blood pressure (mmHg) | 119 | 104-136 | 5,777,980 | 0.2% |
| Diastolic blood pressure (mmHg) | 58 | 50-68 | 5,776,194 | 0.3% |
| Mean blood pressure (mmHg) | 77 | 68-89 | 5,793,250 | 0.2% |
| Body temperature (C) | 37 | 36-38 | 1,733,341 | 0.5% |
| Glucose (mg/dL) | 129 | 107-160 | 1,263,878 | 1.2% |

Under the column 'Missing %' are shown the percentage of admissions for which no measurements were made of the corresponding vital sign.

## Supplementary References